# Select-Mosaic: Data Augmentation Method for Dense Small Object Scenes

Hao Zhang，Shuaijie Zhang，Renbin Zou

**Abstract**

Data augmentation refers to the process of applying a series of transformations or expansions to original data to generate new samples, thereby increasing the diversity and quantity of the data, effectively improving the performance and robustness of models. As a common data augmentation method, Mosaic data augmentation technique stitches multiple images together to increase the diversity and complexity of training data, thereby reducing the risk of overfitting. Although Mosaic data augmentation achieves excellent results in general detection tasks by stitching images together, it still has certain limitations for specific detection tasks. This paper addresses the challenge of detecting a large number of densely distributed small objects in aerial images by proposing the Select-Mosaic data augmentation method, which is improved with a fine-grained region selection strategy. The improved Select-Mosaic method demonstrates superior performance in handling dense small object detection tasks, significantly enhancing the accuracy and stability of detection models. Code is available at https://github.com/malagoutou/Select-Mosaic.

## 1. Introduction

Data augmentation is a common and effective technique in the field of deep learning, which enhances the diversity of training data by generating transformed data, thereby improving the model's generalization ability. With the continuous development of deep learning, data augmentation techniques can be classified into more refined categories, including One-Sample Transform, Multi-Sample Synthesis, and Deep Generative Models.

One-Sample Transform data augmentation is a method focused on generating additional training data by transforming individual samples. Common methods include randomly rotating images by a certain angle, randomly shifting the position of images horizontally or vertically, randomly scaling image sizes, performing horizontal or vertical flips, randomly cropping parts of images, adjusting image brightness, contrast, saturation, etc., adding random noise to images, and applying blur to images, such as Gaussian blur. Random Erase[1], proposed by Zhong et al., randomly selects a rectangular region in the image during training and fills the region with random pixel values (usually gray or the average value of the image).

Multi-Sample Synthesis generates new training data by combining multiple samples, utilizing the diversity of existing data to create richer and more representative training data, thus helping models learn complex features better. Representative methods include Mixup[2] proposed by Zhang et al. in 2018, which generates new training samples by linearly combining two images and their corresponding labels. Yun et al. introduced the CutMix[3] method, which cuts and pastes parts of an image onto another image while adjusting labels to reflect the contribution of the region, increasing data diversity while retaining spatial information.

Mosaic data augmentation technique[4] stitches multiple images together to generate new training samples. It improves upon CutMix by randomly selecting four images from the training set for random combinations. Each image undergoes random scaling and cropping operations, followed by random permutation of the four images to generate a new training sample, thereby enhancing model robustness.

Deep generative models utilize deep learning to generate new data samples. Compared to traditional data augmentation methods, deep generative models can generate more complex and diverse data. Common deep generative models include Generative Adversarial Network[5] (GAN). Deep Generative Models consist of a generator and a discriminator, where the generator generates realistic data samples and the discriminator distinguishes between real and generated data. Through adversarial training, the generator gradually learns to generate high-quality data samples.

## 2.Related Work

CutMix[3] is a data augmentation method employed in image classification tasks aimed at improving the model's understanding and generalization ability towards input images. It combines the ideas of Cutout[6] and Mixup[2]. Firstly, two different images and their corresponding labels are randomly selected. Then, a rectangular region is randomly chosen on one of the images, and the pixel values in this region are replaced with the corresponding pixel values from the same location in the other image. The position and size of this rectangle are randomly determined. Next, the label of the mixed image is computed by taking the weighted average of the labels of the two images based on the proportion of the replaced region. Finally, the model is trained using the mixed image and its corresponding mixed label. CutMix excels in preserving the semantic information of the original images while enhancing the model's robustness to occlusion and noise. By blending content from two different images, the model becomes better equipped to handle various occlusion and noise scenarios, thereby improving its generalization ability. CutMix has been widely applied in various image classification tasks and has shown promising results.

Mosaic data augmentation[4] is a technique used to enhance the performance of object detection models. Its core concept involves synthesizing multiple images into a large mosaic image, which is then used for training. Firstly, four images are randomly selected from the training dataset. Then, a series of random affine transformations are applied to these four images, including scaling, translation, rotation, and shearing, to increase the diversity of the images. Next, the transformed four images are concatenated together in a random order to form a large mosaic image. Subsequently, based on the label information of the original images, the positions and sizes of the objects in the mosaic image are adjusted, and the positions and categories of the objects are re-labeled. Finally, the object detection model is trained using this synthesized mosaic image and its corresponding label information. In this way, Mosaic data augmentation effectively augments the training dataset, enhancing the model's generalization ability and robustness, thereby improving the performance of the object detection model in various scenarios.

## 3. Select-Mosaic

As shown in Figure 1, the schematic diagram of the Select-Mosaic data augmentation method.

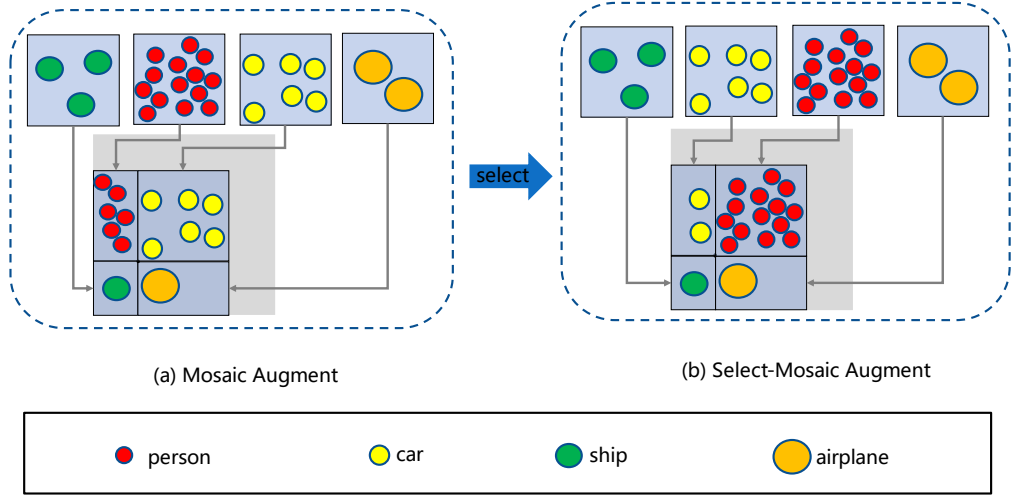

Figure 1: Select-Mosaic Schematic Diagram

Figure (a) depicts the mosaic data augmentation process. Firstly, a mosaic image mask is generated, and then a random coordinate is generated within the mosaic image mask as the splicing center point. Next, four images are randomly selected from the dataset, and freely spliced along the splicing center point within the mosaic image mask. Finally, the mosaic image after data augmentation is obtained.

Figure (b) illustrates the select-mosaic data augmentation process. Before the splicing step, a selection operation is performed. Specifically, specified images are placed in specified positions. Then, selective splicing is conducted. Specifically, the target density within each image is calculated, and the image with the highest target density is identified. The areas of the four masks are calculated, and the mask with the largest area is identified. The image with the highest target density is placed within the mask with the largest area, and the other three images are randomly spliced along the splicing center point within the mosaic mask.

The Select-Mosaic data augmentation steps are illustrated in Figure 2, where (a) represents the Mosaic data augmentation steps, and (b) represents the Select-Mosaic data augmentation steps. Compared to Mosaic data augmentation, Select-Mosaic data augmentation adds the steps "Calculate Target Density & Calculate Mask Area." These steps increase the probability of scenes with dense small objects, making the model more focused on detection in dense scenarios, thereby improving detection performance. During model training, Select-Mosaic is typically used in conjunction with Mosaic. When using Select-Mosaic, the parameter S can be adjusted to control the probability of performing the region selection operation. If a random probability is less than the probability parameter S, Select-Mosaic data augmentation is used, and the region selection operation is executed during data augmentation. If the random probability is greater than the probability parameter S, the region selection operation is not executed, resulting in standard Mosaic data augmentation. This setup aims to control the impact of Select-Mosaic data augmentation on different detection tasks.

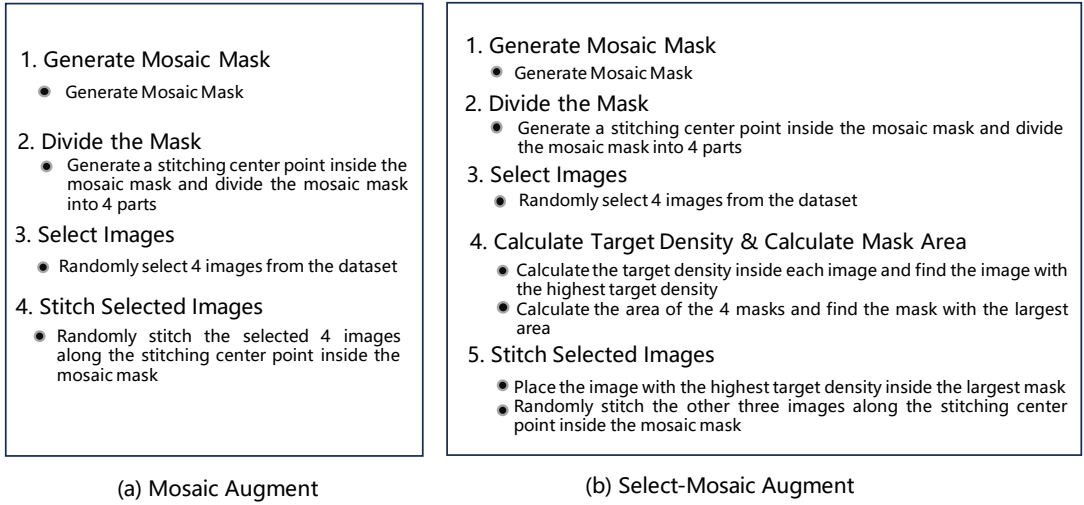

Figure 2: Select-Mosaic Process

## 4.Experiment

To validate the effectiveness of the Select-Mosaic data augmentation described in this paper, experiments were conducted on the AI-TOD remote sensing dataset[9]. YOLOv5[11] was used as the base model, and ablation experiments were performed by setting different values for the region selection probability S. The experimental results are shown in Table 1.

| Methods | P (%) | R (%) | $AP_{50}$(%) | mAP(%) |
|---|---|---|---|---|
| **Mosaic** | 66.65 | 40.30 | 42.70 | 18.06 |
| **Select-Mosaic(S=0.1)** | 67.46 | 41.27 | 43.38 | 18.32 |
| **Select-Mosaic(S=0.2)** | 69.55 | 40.14 | 43.39 | 18.35 |
| **Select-Mosaic(S=0.3)** | 66.54 | 41.57 | 43.09 | 18.39 |
| **Select-Mosaic(S=0.4)** | 67.08 | **41.78** | 44.21 | 18.73 |
| **Select-Mosaic(S=0.5)** | **69.14** | 41.37 | 43.31 | 18.62 |
| **Select-Mosaic(S=0.6)** | 68.33 | 41.14 | 43.70 | 18.66 |
| **Select-Mosaic(S=0.7)** | 67.56 | 41.18 | 43.43 | 18.59 |
| **Select-Mosaic(S=0.8)** | 64.26 | 41.54 | **44.87** | **19.05** |
| **Select-Mosaic(S=0.9)** | 67.22 | 41.72 | 43.84 | 18.64 |
| **Select-Mosaic(S=1.0)** | 66.26 | 41.19 | 42.86 | 18.68 |

Table 1: Ablation Experiment Results for Select-Mosaic Data Augmentation

To demonstrate the generalization capability of Select-Mosaic, comparative experiments were conducted on the Visdrone2019 dataset[10] using YOLOv5 as the base model. The experimental results are shown in Table 2.

| **Methods** | **P (%)** | **R (%)** | **AP$_{50}$(%)** | **mAP(%)** |
|---|---|---|---|---|
| **Mosaic** | 50.22 | **42.93** | 42.73 | 24.86 |
| **Select-Mosaic** | **52.15** | 42.47 | **43.11** | **25.08** |

Table 2: Comparative Experiment Results for Select-Mosaic Data Augmentation

## Conclusion

The Select-Mosaic data augmentation method proposed in this paper introduces a fine-grained region selection strategy, improving the traditional Mosaic data augmentation technique. It is particularly suited for detecting dense small objects in aerial imagery. Experimental results indicate that the enhanced Select-Mosaic method significantly improves the accuracy and stability of detection models when handling dense small object detection. Specifically, compared to Mosaic data augmentation, using Select-Mosaic data augmentation with YOLOv5 resulted in an AP50 improvement of 2.17% on the AI-TOD dataset and 0.37% on the Visdrone2019 dataset. In summary, the Select-Mosaic data augmentation method proposed in this paper effectively addresses the limitations of the traditional Mosaic method in dense small object detection tasks through a fine-grained region selection strategy. Future research could further explore the applicability and extensibility of this method in other types of object detection tasks, thereby advancing data augmentation techniques in the field of computer vision.